# Weight-Space Programming Changes How an LLM Uses Python for Computational Biology

Makoto Sato

Mathematical Neuroscience Unit, Institute for Frontier Science Initiative, Laboratory of Developmental Neurobiology, Graduate School of Medical Sciences, Kanazawa University, Kanazawa, Ishikawa, Japan.

Correspondence: makotos@staff.kanazawa-u.ac.jp

## Abstract

Weight-space programming allows a language model to edit another model's weights without gradient-based training. We used Claude Opus 5.5 to edit Qwen3.5-0.8B and investigated how the edits changed its use of Python on computational biology problems. We compared four initial edited checkpoints with the original model on 50 test problems across three test seeds, yielding 600 paired evaluations. Strictly correct answers increased from 44 to 77, and technically correct answers, which included correct final values rejected for formatting, increased from 64 to 100. Repeated Python calls decreased from 1,834 to 1,044. Among the 64 pairs in which only the edited response was technically correct, the original model had already computed the correct value in 36 but did not submit a technically correct final answer. The edited model corrected the computation in 13 of the 64 pairs. With six additional checkpoints on the same 50 problems, technically correct answers rose from 96 to 154 across 900 pairs. We explored three selected problems with ten test seeds for each of ten checkpoints. On the melting-temperature problem, correct Python outputs decreased from 60 to 22 while technically correct final answers increased from 0 to 18. Weight editing changed both the calculations Qwen ran and whether it submitted a correct answer.

## Introduction

Neural networks typically acquire their weights through gradient-based training. Identifying which computations support a particular behavior and changing that behavior selectively remain difficult. Mechanistic interpretability has provided ways to trace selected computations through transformer components [1], and model-editing methods have shown that direct weight updates can alter factual associations [2].

Takehi and colleagues introduced weight-space programming, in which a general-purpose language model inspects and modifies network weights to pursue a goal specified in natural language [3]. Their demonstrations included repairing a game-playing policy and constructing policies for other games. In a separate experiment, they edited Qwen3.5-0.8B with GPT-6 Astra and reported less repetitive reasoning and improved scores on mathematics and instruction-following benchmarks. Hu and colleagues subsequently evaluated WSP on Qwen3.5-0.8B and Qwen3.5-9B and reported substantial variation across editing runs [4]. These findings raise a question about how an edit improves problem solving. It may help a model finish a solution it can already compute, or it may change the calculation the model performs. A benchmark score alone leaves these possibilities unresolved.

Program-aided language models have shown how generated code can delegate computation to a Python interpreter [5]. We study weight editing on computational biology problems where Qwen can execute Python. The executed code and its output show how the model computed a result, and the final answer shows what it submitted. We used Claude Opus 5.5 to edit Qwen3.5-0.8B and examined these records in detail for four initial checkpoints. Pairing each checkpoint with the original model on 50 test problems withheld from the Editor at three test seeds yielded 600 comparisons (4 × 3 × 50, Fig. 1). We compared correct Python outputs with correct final answers to examine how computation and submission varied across trials.

## Methods

### Experimental overview

Figure 1 summarizes the workflow. Each editing run had a target-call budget (TCB) of up to 10 successful calls. The first call evaluated the original Qwen on all 50 editing problems. Claude Opus 5.5 inspected Qwen's responses and used the editing problems, reference answers, and step-by-step explanations to guide direct edits to Qwen's weights. At the end of each run, we froze the edited checkpoint after confirming that it answered more editing problems correctly than the original model. In the initial evaluation, we tested both models on the same 50 test problems. Both versions could execute Python. We compared their code, execution outputs, termination outcomes, and final answers.

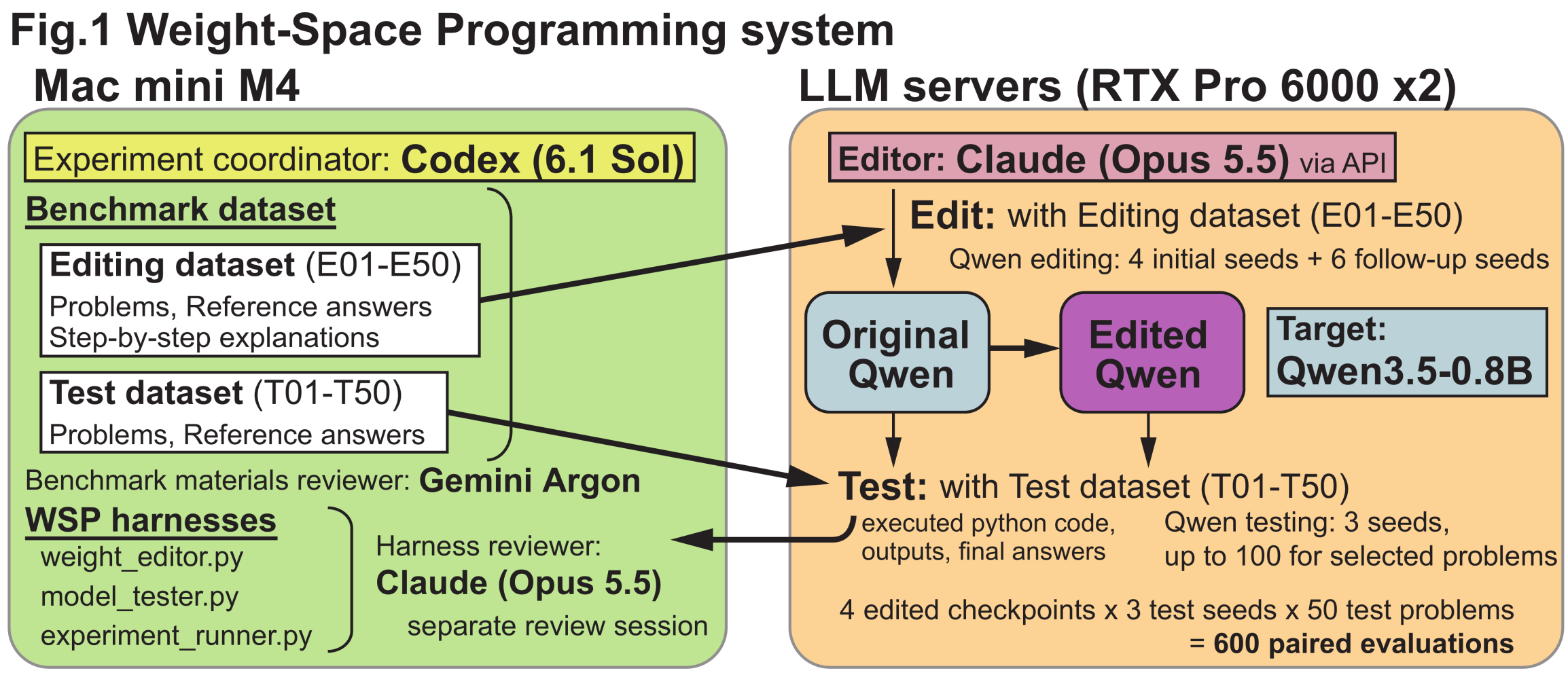

**Figure 1.** Weight-space programming system and paired evaluation. Claude Opus 5.5 was accessed remotely via API. Qwen ran on two LLM servers. The accepted edited checkpoints were frozen before Test. Both Qwen versions received the test problems, and the test reference answers were reserved for scoring. For T03, testing extended through seed 100 in selected conditions (Supplementary Table S1).

## Computational biology problems and step-by-step explanations

The benchmark dataset comprised two sets of 50 computational biology problems: an editing dataset used during weight editing (E01-E50) and a test dataset withheld from the Editor during weight editing (T01-T50). Earlier versions drew on sequence-analysis tasks and validators from LABBench2 SeqQA2 [6]. The test dataset combined 20 existing problems and 30 newly generated instances of related computational tasks. The editing dataset also included newly constructed primer-interaction problems. Topics included molecular mass, melting temperature, sequence entropy, enzyme kinetics, hydrophobicity, and codon optimization. We specified the quantities to compute, units, relevant conventions, and input files. The step-by-step explanations for the editing dataset were initially drafted with GPT-6.1 Sol. We checked them against reference calculations and revised them as we updated the problems and grading rules.

During development, Gemini Argon independently solved all 100 problems in an earlier version of the benchmark before seeing the reference materials. It then compared its answers with the reference answers and step-by-step explanations and reviewed the grading code. We checked the final benchmark problems and grading rules against local reference calculations. The Editor, Claude Opus 5.5, received the editing dataset, including the problems, reference answers, and step-by-step explanations. The test dataset was withheld from the Editor, and its reference answers were reserved for scoring.

## Editing and Python execution harnesses

Claude Opus 5.5 served as the Editor and Qwen3.5-0.8B as the Target [7]. Qwen ran locally with BF16 weights and thinking enabled, using vLLM [8] on two servers equipped with NVIDIA RTX PRO 6000 GPUs. Each editing run began with the original checkpoint and a fresh Editor session. We used a different Qwen sampling seed for each initial editing run (seeds 1–4) and each follow-up editing run (seeds 5–10). The Editor could inspect tensors, execute calculations in an isolated CPU environment, modify weights through WSP harnesses, and evaluate candidate behavior on problems in the editing dataset. It performed direct weight updates without gradient-based training. A single target call could evaluate multiple problems. In addition to the TCB, each editing run was limited to 32,768 Editor output tokens, 300 Editor CPU calls, and six hours.

We accepted a final candidate only if its completed-correct count exceeded that of the original model in paired evaluations of all 50 problems in the editing dataset. We evaluated the final candidates twice for editing seeds 1 and 2. The original-to-edited counts were 4 to 11 for seed 1 and 1 to 8 for seed 2 in both confirmations, so the second confirmation did not change the acceptance decision. We evaluated the final candidates once for seeds 3 and 4, with counts of 5 to 10 and 6 to 7, respectively. The accepted checkpoints remained fixed throughout evaluation on the test dataset. During development, Claude Opus 5.5 reviewed the WSP harnesses used for editing and evaluation in a separate session.

Qwen could request Python execution in a separate isolated process for each problem. The harness returned outputs and ordinary execution errors to Qwen and recorded the code, standard output and error, and terminal status. Both original and edited versions had access to the same input files and Python environment, which included NumPy, the Primer3 Python bindings [9], and Biopython [10]. For each test problem, Qwen could generate up to 32,768 tokens, make up to 12 Python calls, and take up to 24 turns. Sampling used a temperature of 0.6, top-p of 0.95, top-k of 20, and a presence penalty of 1.5. A response could terminate upon submission, a budget limit, or an execution failure.

## Paired testing

For each of the four initial frozen candidates, we evaluated the original and edited checkpoints on the same server using the same 50 test problems and three sampling seeds (1–3). This yielded 150 paired evaluations per candidate and 600 in total, comprising 1,200 response records. In this initial evaluation, the same 50 problems recurred across the four candidates and three test seeds. The original checkpoint was evaluated twice for each server, problem, and test seed combination. Across these 300 combinations, both original evaluations had identical executed code, standard output and error, final text, and termination reason in 286 cases. We summarized the paired evaluation outcomes using descriptive counts. For the problem-level analysis, we pooled the 12 pairs per problem across the four editing candidates and three test seeds.

For the full-dataset follow-up, we evaluated six additional checkpoints (editing seeds 5–10) using the same 50 problems and test seeds 1–3. This added 900 pairs (Table 5). The editing run for seed 6 stopped when the weight-saving tool received an invalid argument under WSP harness v4.1. We started a new run from the original checkpoint with a fresh Editor session and WSP harness v4.2, which changed how the tool handled invalid arguments. We evaluated this checkpoint under the same full-dataset test protocol (Table 5) and in the selected-problem follow-up (Table 6).

For T02, T17, and T03, we compared ten test seeds for each editing seed. For the nine v4.1 checkpoints, we used the corresponding trials from the full 50-problem evaluation for test seeds 1–3 and ran each problem individually for seeds 4–10. We matched Original and Edited on server, problem, and test seed. We reused the same standalone Original response for edited checkpoints evaluated on the same server for that problem and seed. Table 6 therefore reports descriptive counts across 100 matched pairs per problem.

The automated grader counted a response as strictly correct if the model submitted a final answer in the required format and met the correctness criteria for that problem. We used the same grading rules for the original and edited models. For the secondary count, we retrospectively reviewed the final text of responses rejected for formatting using records labeled by model condition. The technically correct count included all strictly correct responses plus responses rejected for formatting whose final text explicitly stated the requested value and matched the reference answer within the numerical tolerance for that problem.

For each response, we parsed the code from each Python call with ‘ast.parse’ and serialized its abstract syntax tree (AST) with ‘ast.dump(include_attributes=False)’, omitting source location information. If parsing failed, we used the exact source text for matching. We counted each occurrence of the same representation after the first as a repeated call, even when the matching calls were nonconsecutive. This measure can miss equivalent computations written with different syntax.

**Analysis of executed code and outputs**

In the initial evaluation of editing seeds 1–4, Codex (GPT-6.1 Sol) compared the original and edited responses using the saved Python execution logs. For each pair, it checked whether the executed code used the input data, calculation, and units specified in the problem, whether the recorded output matched the reference result, and whether the model reported that result in its final answer. A calculation was classified as corrected when the edited code fixed a specific error in the original computation and the execution log confirmed the expected result. Input-handling fixes, execution errors, fewer recalculations, and successful submission of an already correct result were recorded separately.

When the edited model did not execute Python, Codex used a shorter review. It checked whether the original model had computed the requested result, noted the main failure observed, and examined the edited model's termination status and any final answer. When the logs provided too little evidence to determine whether the calculation was correct, Codex recorded the judgment as undetermined.

For the full-dataset follow-up, we summarized automated grades, completion, Python calls, and repeated calls, and reviewed final responses rejected for formatting. We inspected selected execution logs from those checkpoints. For the three problems in Table 6, we also checked whether any Python call produced the correct result from the specified input. We counted a correct Python output once per response, even if the model did not submit it. Strictly and technically correct counts used the final answer.

# Results

## Test performance across editing runs

Across 600 paired evaluations, the number of responses graded as strictly correct increased from 44 (7.3%) for the original model to 77 (12.8%) for the edited models (Fig. 2A and Table 1). Strict correctness required both a correct answer and the prescribed submission format. The technically correct count included all strictly correct responses and also counted responses whose final text stated the requested value within the numerical tolerance for the problem but failed the format requirements. By this criterion, the number of correct responses increased from 64 (10.7%) to 100 (16.7%), with an increase in each of the four editing runs. We use technical correctness in the analyses below to assess whether the model reported the correct result in its final response.

For each problem, we compared the number of technically correct responses from the original and edited models across 12 paired evaluations. After editing, this count increased for 17 of the 50 problems, stayed the same for 27, and decreased for six. Under strict grading, the corresponding counts were 18, 27, and five. Unfinished responses decreased from 507 to 347, while completed responses that did not meet the technical correctness criterion increased from 29 to 153 (Fig. 2B). Terminations at the Python call limit decreased from 398 to 205, while terminations at the generation budget increased from 97 to 137.

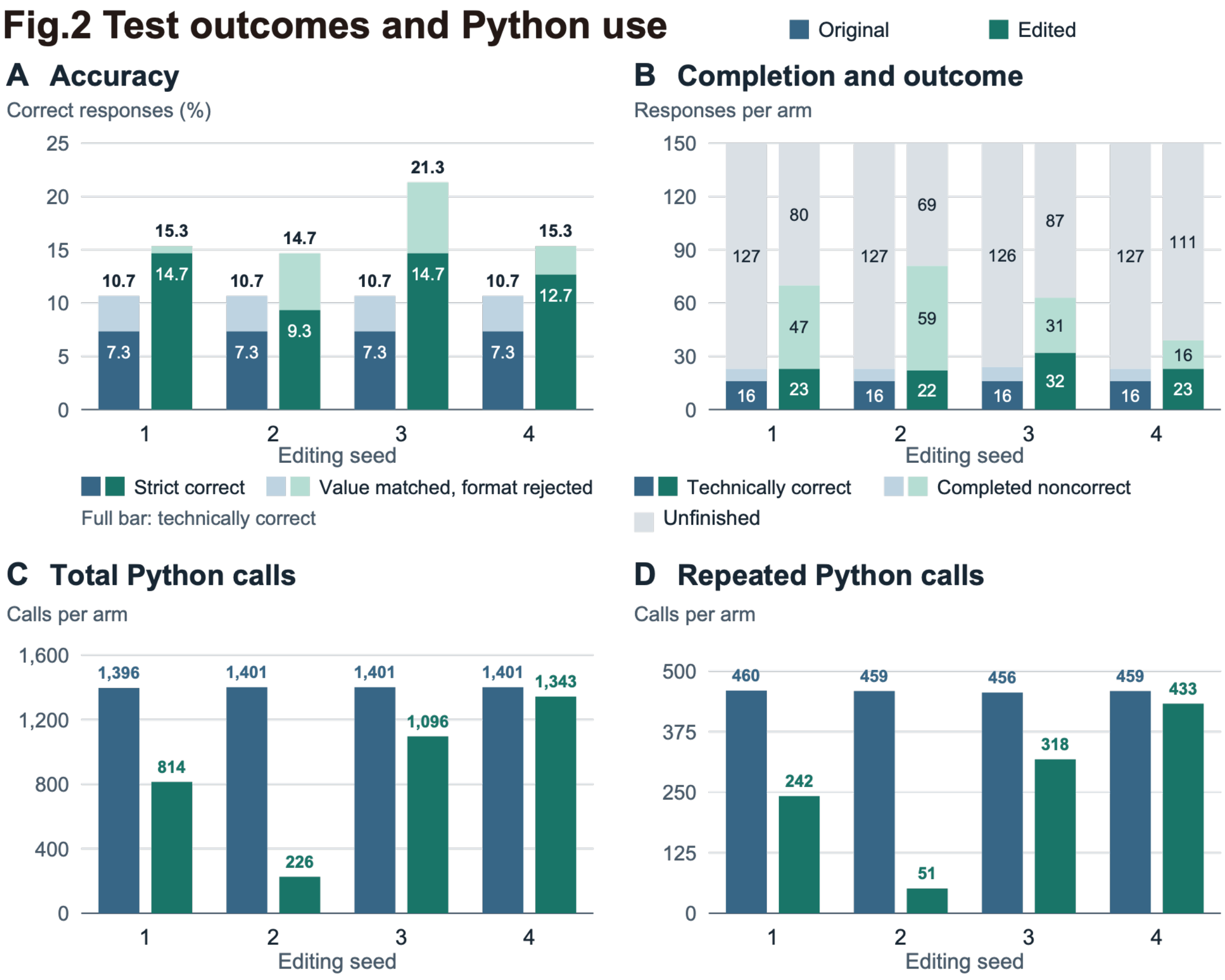

**Figure 2.** Test accuracy, completion, and Python use by editing seed. Each condition contributed 150 responses per editing seed. (A) The dark segments show strict accuracy. The full bars also count final values that matched the reference within tolerance but failed the format requirements. (B) Counts of technically correct, completed noncorrect, and unfinished responses. “Completed noncorrect” denotes responses that reached a final answer but did not meet the criteria for technical correctness. (C) Total Python calls. (D) Number of repeated Python calls. A call was counted as repeated if its code had the same abstract syntax tree as an earlier call in the same response. If parsing failed, we matched the source text exactly.

**Table 1.** Paired Test performance by editing seed. Arrows run from Original to Edited.

| Editing seed | Pairs | Strict correct | Technically correct | Gains / losses | Unfinished |
|---|---|---|---|---|---|
| 1 | 150 | 11 (7.3%) → 22 (14.7%) | 16 (10.7%) → 23 (15.3%) | 16 (10.7%) / 5 (3.3%) | 127 (84.7%) → 80 (53.3%) |
| 2 | 150 | 11 (7.3%) → 14 (9.3%) | 16 (10.7%) → 22 (14.7%) | 12 (8.0%) / 9 (6.0%) | 127 (84.7%) → 69 (46.0%) |
| 3 | 150 | 11 (7.3%) → 22 (14.7%) | 16 (10.7%) → 32 (21.3%) | 16 (10.7%) / 5 (3.3%) | 126 (84.0%) → 87 (58.0%) |
| 4 | 150 | 11 (7.3%) → 19 (12.7%) | 16 (10.7%) → 23 (15.3%) | 14 (9.3%) / 6 (4.0%) | 127 (84.7%) → 111 (74.0%) |
| All | 600 | 44 (7.3%) → 77 (12.8%) | 64 (10.7%) → 100 (16.7%) | 58 (9.7%) / 25 (4.2%) | 507 (84.5%) → 347 (57.8%) |

Technically correct adds format-rejected final responses that stated the requested value. Gains and losses use strict grading. Percentages use the number of pairs in each row as the denominator.

All four retained checkpoints had changes in the MLP down-projection of the final Transformer block (Table 2). For editing seeds 1, 3, and 4, changes were confined to that matrix. Seed 2 showed changes in 24 matrices spanning blocks 12–23.

**Table 2.** Weight changes in the retained checkpoints.

| Editing seed | Changed weights | Block indices | Matrices with changes |
|---|---|---|---|
| 1 | 64,409 | 23 | 1 MLP down-projection |
| 2 | 51,892,555 | 12–23 | 12 MLP down-projections and 12 attention output projections |
| 3 | 616,799 | 23 | 1 MLP down-projection |
| 4 | 1,019 | 23 | 1 MLP down-projection |

Block indices are zero-based. Counts include scalar weight values that differed from the original checkpoint at BF16 precision.

## Repeated Python execution

Total Python calls fell from 5,599 to 3,479 after editing, and calls repeating an earlier syntax tree fell from 1,834 to 1,044 (Fig. 2C, D and Table 3). The fraction of calls that repeated an earlier syntax tree decreased from 32.8% to 30.0%. Repetition decreased in 295 pairs, was unchanged in 161, and increased in 144. The size of the reduction varied across editing runs. For editing seed 2, repeated calls fell from 459 to 51. For editing seed 4, they fell from 459 to 433. In T27 (editing seed 3, test seed 1), the original model obtained the correct rate of 3 $s^{-1}$ and kept recalculating until it reached the tool limit. The edited model computed the same value and submitted it after three Python calls (Fig. 3A).

Across the 600 paired evaluations of 50 problems using editing seeds 1–4 and test seeds 1–3, 20 edited responses were technically correct without calling Python. No original response was technically correct without calling Python. For example, on T29 (editing seed 1, test seed 2), the edited model calculated and submitted the correct turnover number of 3.5 $s^{-1}$ without using Python.

**Table 3.** Python calls and repeated syntax within responses.

| Editing seed | Python calls | Repeated calls (% of Python calls) |
|---|---|---|
| 1 | 1,396 → 814 | 460 (33.0%) → 242 (29.7%) |
| 2 | 1,401 → 226 | 459 (32.8%) → 51 (22.6%) |
| 3 | 1,401 → 1,096 | 456 (32.5%) → 318 (29.0%) |
| 4 | 1,401 → 1,343 | 459 (32.8%) → 433 (32.2%) |
| All | 5,599 → 3,479 | 1,834 (32.8%) → 1,044 (30.0%) |

## Changes in the computations performed by Python

Of the 64 pairs in which only the edited response was technically correct, 13 showed a correction to the target computation in the executed Python code. In 36 pairs, the original model had already computed the correct value in Python. In 9 of the 13 pairs with a corrected computation, the edited response also had fewer repeated Python calls.

In T02 (editing seed 1, test seed 2), the original model used raw nucleotide counts in the Shannon entropy formula and obtained −46.43856. The edited model normalized the counts by dividing each by the sequence length, then computed Shannon entropy from the resulting probabilities. It obtained and submitted the correct value of 2 bits (Fig. 3B, Table 4).

In T17 (editing seed 3, test seed 2), the edited model used log10 in the salt-correction term, computed the correct melting temperature of 31.4 °C, and submitted that value (Fig. 3C, Table 4). The explanation listed 26 bases instead of 16, but the calculation used the correct nucleotide counts.

In T03 (editing seed 2, test seed 1), the original model encountered Python errors while assigning codons and left the answer unfinished. The edited model selected the maximum-weight synonymous codon for each amino acid position, joined the 99 codons, and submitted the resulting 297-nucleotide DNA sequence. The automated grader accepted the sequence (Fig. 3D). This was the only correct final answer for this checkpoint across test seeds 1–100 (Supplementary Table S1).

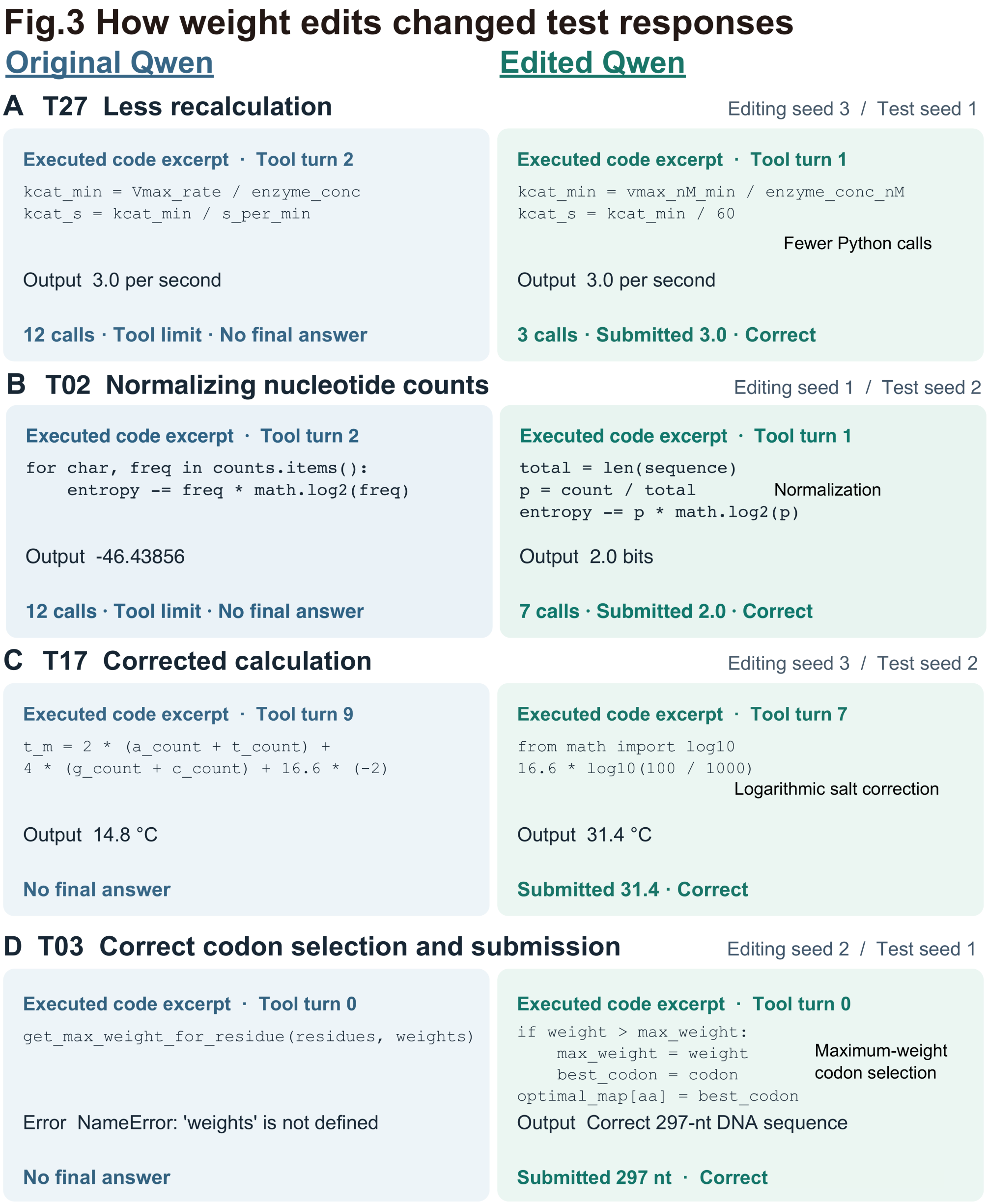

**Figure 3.** Representative Python execution traces before and after weight editing. (A) Less recalculation (T27, editing seed 3, test seed 1). Both models correctly computed a turnover number of 3.0 $s^{-1}$. The original model continued recalculating and reached the 12-call limit without submitting a final answer. The edited model submitted the correct value after three Python calls. (B) Normalization of nucleotide counts (T02, editing seed 1, test seed 2). The original model used raw nucleotide counts in the Shannon entropy formula and obtained −46.43856. The edited model normalized each count by dividing it by the sequence length, then used the resulting probabilities to compute and submit the correct entropy of 2 bits. (C) Corrected salt-concentration calculation (T17, editing seed 3, test seed 2). The original model used −2 for the logarithm in the salt correction and obtained 14.8 °C. The edited model converted 100 mM to 0.1 M and evaluated log10(100 / 1000) to calculate the salt correction. It computed and submitted the correct melting temperature of 31.4 °C. (D) Correct codon selection and submission (T03, editing seed 2, test seed 1). The original model encountered Python errors and did not submit a final answer. The edited model selected the maximum-weight synonymous codon at each of the 99 amino acid positions and joined the codons into a 297-nucleotide DNA sequence. It submitted the same sequence as its final answer, which passed automated grading.

In T47 (editing seed 1, test seed 1), the original model's regular expression extracted the FASTA header identifier instead of the amino acid sequence, causing a 'KeyError' during the hydrophobicity lookup. The edited model specified the correct sequence directly in its Python code, bypassing FASTA parsing (Table 4). It computed the minimum mean hydrophobicity across all five-residue windows (−2.9) and reported that value in its final response. The automated grader rejected the response because it lacked the required answer tag, but we counted it as technically correct.

**Table 4.** Selected paired execution traces. Seed pairs give editing seed followed by test seed.

| Task and seeds | Original execution and answer | Edited execution and answer |
|---|---|---|
| T27 · 3 / 1 | Computed 3 $s^{-1}$. Reached the tool limit without submitting. | Computed and submitted 3 $s^{-1}$ after three calls. |
| T02 · 1 / 2 | Used nucleotide counts in the entropy formula. Printed −46.43856 and left the response unfinished. | Normalized counts to probabilities. Computed and submitted 2 bits. |
| T17 · 3 / 2 | Substituted −2 for the logarithmic term. Printed 14.8 °C and left the response unfinished. | Used log10 in the salt correction. Computed and submitted 31.4 °C. |
| T47 · 1 / 1 | Treated the FASTA header as sequence. Raised a KeyError and did not submit. | Specified the correct sequence directly in the code and computed the minimum window mean (−2.9). Reported the value without the required answer tag. |
| T03 · 3 / 1 | Encountered code errors and did not produce a valid codon sequence. | Produced a valid codon sequence. Failed its own verification and did not submit. |
| T20 · 3 / 3 | Computed 64 °C and left the response unfinished. | Miscounted bases and submitted 32 °C. |

## Remaining failures and regressions

A corrected computation could still fail during verification or submission. In T03 (editing seed 3, test seed 1), the edited model executed code that generated a sequence satisfying the stated codon-weight criterion and translating to the required protein. Its self-check compared amino acid symbols with DNA codons, and the model did not submit a final answer (Table 4).

Changes could also degrade the computation. In T20 (editing seed 3, test seed 3), the original model computed the correct melting temperature of 64 °C and left the answer unfinished. The edited model miscounted the bases and submitted 32 °C (Table 4). Neither response counted as technically correct.

## Follow-up evaluations

Across the six additional checkpoints, technically correct responses increased from 96 to 154 in 900 pairs on the full test dataset (Table 5). We also evaluated three selected problems across ten test seeds per editing seed. On T17, the number of correct Python outputs decreased from 60 to 22, while the number of technically correct final responses increased from 0 to 18 (Table 6).

**Table 5.** Follow-up performance on the 50-problem test dataset.

| Editing seed | Pairs | Strictly correct | Technically correct | Unfinished | Completed noncorrect | Python calls | Repeated calls |
|---|---|---|---|---|---|---|---|
| **Initial (1–4)** | 600 | 44→77 | 64→100 | 507→347 | 29→153 | 5,599→3,479 | 1,834→1,044 |
| **Follow-up (5–10)** | 900 | 66→129 | 96→154 | 762→472 | 42→274 | 8,393→3,897 | 2,733→1,154 |
| **All 10 candidates** | 1,500 | 110→206 | 160→254 | 1,269→819 | 71→427 | 13,992→7,376 | 4,567→2,198 |

Values are Original → Edited. Each candidate was evaluated on 50 test problems at test seeds 1–3, giving 150 pairs. The follow-up candidates were editing seeds 5–10. Only editing seed 6 used the v4.2 editing harness. Completed noncorrect responses were completed responses that did not meet the technical correctness criterion. Repeated calls were identified by matching Python syntax trees within each response, with exact text matching when parsing failed.

**Table 6.** Ten-seed comparison for T02, T17, and T03.

| Editing seed | | T02 | | | T17 | | | T03 | |
|---|---|---|---|---|---|---|---|---|---|
| | Output | Strict | Technical | Output | Strict | Technical | Output | Strict | Technical |
| 1 | 3→5 | 1→2 | 1→2 | 6→1 | 0→1 | 0→1 | 1→0 | 0→0 | 0→0 |
| 2 | 3→0 | 1→1 | 1→1 | 6→0 | 0→0 | 0→1 | 1→1 | 0→1 | 0→1 |
| 3 | 3→4 | 1→3 | 1→3 | 6→5 | 0→3 | 0→4 | 1→2 | 0→0 | 0→0 |
| 4 | 3→6 | 1→0 | 1→0 | 6→2 | 0→0 | 0→0 | 1→1 | 0→0 | 0→0 |
| 5 | 3→3 | 1→3 | 1→3 | 6→4 | 0→2 | 0→2 | 1→1 | 0→0 | 0→0 |
| 6 | 3→4 | 1→2 | 1→2 | 6→4 | 0→3 | 0→4 | 1→0 | 0→0 | 0→0 |
| 7 | 3→0 | 1→0 | 1→1 | 6→0 | 0→0 | 0→0 | 1→0 | 0→0 | 0→0 |
| 8 | 3→3 | 1→1 | 1→1 | 6→1 | 0→2 | 0→2 | 1→0 | 0→0 | 0→0 |
| 9 | 3→5 | 1→1 | 1→2 | 6→4 | 0→2 | 0→2 | 1→1 | 0→0 | 0→0 |
| 10 | 4→0 | 1→0 | 1→0 | 6→1 | 0→2 | 0→2 | 1→0 | 0→0 | 0→0 |
| **Total** | **31→30** | **10→13** | **10→15** | **60→22** | **0→15** | **0→18** | **10→6** | **0→1** | **0→1** |

Values are Original → Edited. Output counts trials in which Python produced the correct result. For T03, a correct Python output had to encode the full target protein and use a synonymous codon with the highest supplied weight at each position. Strict and Technical score final responses. Each editing seed contributed ten pairs per problem. Editing seed 6 used the v4.2 editing harness and standalone trials for seeds 1–10. Other editing seeds used the 50-problem evaluation for seeds 1–3 and standalone trials for seeds 4–10. Standalone Original responses were reused across edited checkpoints on the same host.

# Discussion

## Weight edits changed the computations executed by the model

Takehi and colleagues [3] reported that editing Qwen reduced repeated reasoning and helped the model finish answers it already knew how to compute. We observed a similar pattern in T27 (Fig. 3A). Both models computed the correct turnover number of 3.0 $s^{-1}$. The original model continued recalculating until it reached the Python-call limit, while the edited model submitted the value after three calls. Here, the difference was whether the model used the result it had already obtained to finish its answer.

In T02, the executed calculation itself changed (Fig. 3B). The original model inserted raw nucleotide counts into the Shannon entropy formula and obtained −46.43856. The edited model divided each count by the sequence length to obtain probabilities and used them to compute and submit the correct entropy of 2 bits. Building on Takehi and colleagues' analysis of reasoning repetition [3], we used the saved Python code, execution output, and final response to trace how weight editing changed the computational procedure or the use of its result in individual problems.

In the full-dataset follow-up, technically correct responses increased from 96 to 154 and unfinished responses fell from 762 to 472 across 900 pairs (Table 5). Completed responses that failed the technical correctness criterion also increased, from 42 to 274. The additional checkpoints thus produced more completed responses in both outcome categories.

## Changes in the selection of a solution

In the T17 example (Fig. 3C, editing seed 3, test seed 2), the original model used −2 for the logarithm in the salt correction and computed 14.8 °C. The edited model converted 100 mM to 0.1 M and evaluated log10(0.1) to obtain the correct melting temperature of 31.4 °C, which it submitted. In another trial with test seed 1, the original model used log10 correctly and printed 31.4 °C, but reached the generation limit before submitting a final answer. Across all ten editing seeds and ten test seeds (Table 6), correct Python outputs decreased from 60 to 22 after editing, while strictly correct final answers increased from 0 to 15 and technically correct final answers from 0 to 18. Of the 18 technically correct edited responses, 13 had a correct Python output, three made no Python calls, and two called Python without producing the correct output. In T17, correct Python outputs became less frequent after editing. Nevertheless, the edited models produced more correct final answers, including several for which the execution logs showed no correct Python output.

In the broader comparison for T02, the original model also produced the correct entropy, and the number of trials with correct Python output was similar before and after editing (31 versus 30, Table 6). The technically correct final-answer count increased from 10 to 15. Together, T02 and T17 show that an edited checkpoint can use a valid procedure in a particular trial even when the frequency of correct output does not increase across the tested seeds.

T03 showed how verification could prevent submission of a correct result. In a standalone trial at test seed 9, the original model produced a valid codon sequence and left the answer unfinished. With editing seed 2 and test seed 1, the edited model generated and submitted the correct 297-nucleotide sequence (Fig. 3D). With editing seed 3 and test seed 1, it also generated a valid sequence, but a self-check compared amino acid symbols with DNA codons, and the model left the answer unfinished (Table 4). The execution logs therefore distinguish successful codon selection from successful verification and submission.

A correct value can also be rejected at grading. In T47, the edited model reported the correct value in its final response. The automated grader rejected the response because it lacked the required answer tag, and we counted it as technically correct (Table 4).

## Scope and limitations

We studied one Target model and one Editor model. We evaluated all ten checkpoints on the same 50 test problems using three sampling seeds. The test problems were known to the study author, and some were revised during benchmark development after earlier evaluations. Some test problems required the same computational procedures as problems in the editing dataset, using different inputs. The magnitude of the improvement varied across editing runs, as Hu and colleagues [4] also observed.

Counting repeated Python syntax can miss equivalent calculations written in different ways (Fig. 2D). Detailed comparison of execution traces for the four initial candidates revealed changes in individual cases. In the follow-up evaluation of all 50 problems, we analyzed grading outcomes and tool use for the six additional candidates (editing seeds 5–10). We reviewed only selected execution traces and therefore did not estimate an overall calculation repair rate for these candidates.

## Linking weight changes to executed behavior

The MLP down-projection matrix in the final Transformer block changed in all four initial checkpoints and was the only matrix that changed in three of them (Table 2). Hu and colleagues [4] modified the same matrix in all three of their mathematics editing runs on Qwen3.5-0.8B. The matrix was therefore a common target across two task domains. One possible reason is that changes to the final-block MLP output reach the model's token readout without passing through further Transformer blocks, which may make their immediate effects on next-token predictions easier to estimate. Why the Editor repeatedly selected this matrix remains an open question.

Takehi and colleagues [3] reduced reasoning repetition in the same model by editing MLP weights in earlier blocks, layers 15 and 19. In layer deletion and swapping experiments, Lad and colleagues [11] found that next-token prediction was sensitive to interventions near the end of the model and proposed that the final block helps refine the output distribution. Stolfo and colleagues [12] showed that individual neurons in the final-block MLP can influence next-token confidence, including when the model continues a repeated sequence. These findings motivate testing whether the layer-23 edits contributed to the changes in Python use and answer submission observed here.

A next step is to focus on editing seeds 1, 3, and 4, for which all weight changes were confined to the MLP down-projection in layer 23. For each checkpoint, we could restore subsets of the changed weights to their original values and separately apply those same changes to the original checkpoint. Using the same test problems and sampling seeds, we could then assess which subsets contribute to changes in the computations performed in Python and in whether the model submits the computed result.

A related question is how the Editor chose its weight updates. The Editor may have used changes in the Target model's performance after small weight perturbations to infer which directions to try next. To test this possibility, future work could identify which proposed edits were evaluated and examine whether the resulting feedback informed the direction of subsequent updates.

## Concluding remarks

Direct weight edits changed how Qwen used Python to solve computational biology problems. In some cases, the edited model submitted a correct result after fewer recalculations. In others, it changed the calculation itself. Comparing the executed code, its output, and the final answer revealed where these changes occurred.

Weight-space programming is still in its early stages. Future studies could test the approach on a wider range of models and tasks, investigate how specific weight edits change model behavior, and explore practical applications. By showing how to track changes in model behavior from code execution to final answers, this study lays an experimental foundation for the further development of weight-space programming.

## Funding

This work was supported by JSPS KAKENHI under grant numbers 22H05169, 22H05621, 24H00188, 24H01396, and 25K02282.

## Author contributions

M.S. conceived and designed the study, implemented the code, performed the computational experiments, analyzed the data, interpreted the results, and wrote the manuscript.

## Code availability statement

The code developed in this study will be available online (https://github.com/satouma7/bioWSP).

## Conflicts of Interest

The author declares no conflicts of interest associated with this manuscript.

**Supplementary Table S1. Extended evaluation of T03 under standard test budgets.**

| Evaluation | Pairs | Correct Python output | Strictly correct | Technically correct |
|---|---|---|---|---|
| All 10 checkpoints, test seeds 11–20 | 100 | 0 → 4 | 0 → 0 | 0 → 0 |
| Editing seed 2, test seeds 21–100 | 80 | 6 → 3 | 0 → 0 | 0 → 0 |
| Editing seed 2, test seeds 1–100 (cumulative) | 100 | 7 → 4 | 0 → 1 | 0 → 1 |

Counts are Original → Edited. The 100 pairs for test seeds 11–20 reuse 20 Original runs matched by server and test seed. Editing seed 6 used the v4.2 editing harness. The cumulative row includes the editing-seed-2 pairs from test seeds 11–20 and all 80 pairs from test seeds 21–100. For editing seed 2, test seeds 1–3 came from full-dataset evaluation and test seeds 4–100 came from standalone runs. The standard budget was 32,768 generated tokens, 12 Python calls, and 24 turns.